\documentclass[letterpaper,twocolumn,10pt]{article}
\usepackage{usenix-2020-09}

\usepackage{amsmath,amssymb}
\usepackage{graphicx}
\usepackage{placeins}
\usepackage{stfloats}
\usepackage{booktabs}
\usepackage{comment}
\usepackage{xurl}

\hypersetup{
  hidelinks,
  pdftitle={ServerlessLoRA: Enabling Low-Latency Serverless Multi-LoRA Serving},
  pdfauthor={Yifan Sui, Hao Wang, Hanfei Yu, Kaiqiang Xu, Yitao Hu, Chen Chen, Jianxun Li, Kai Chen},
  pdfsubject={Serverless multi-LoRA serving}
}

\usepackage{tikz}
\usepackage{makecell}
\usepackage{subcaption} 
\usepackage{listings}

\definecolor{dkgreen}{rgb}{0,0.6,0}
\definecolor{gray}{rgb}{0.5,0.5,0.5}
\definecolor{mauve}{rgb}{0.58,0,0.82}
\newcommand{\sys}{\mbox{\textit{ServerlessLoRA}}\xspace}

\usepackage{bbold}
\usepackage{multirow}
\usepackage{enumitem}
\usepackage{soul}
\usepackage{color}
\usepackage{acronym}
\usepackage{xspace}
\usepackage[normalem]{ulem}
\usepackage{tabularx}
\usepackage{multirow}
\usepackage[ruled,vlined]{algorithm2e}
\usepackage[usestackEOL]{stackengine}

\definecolor{ForestGreen}{RGB}{34,139,34}

\DeclareRobustCommand{\wdcut}[1]{}

\usepackage{fancyvrb}

\makeatletter
\renewcommand\paragraph{\@startsection{paragraph}{4}{\z@}%
  {4pt plus 1pt minus 1pt}{-1em}{\normalfont\normalsize\bfseries}}
\makeatother

\definecolor{amaranth}{rgb}{0.9, 0.17, 0.31}

\SetKwInput{KwInput}{{\textbf{Input}}}
\SetKwInput{KwOutput}{{\textbf{Output}}}
\SetKwIF{If}{ElseIf}{Else}{{\textbf{if}}}{{\textbf{then}}}{{\textbf{else if}}}{{\textbf{else}}}{{\textbf{endif}}}
\SetKwFor{For}{{\textbf{for}}}{{\textbf{do}}}{
{\textbf{endfor}}}
\SetKwFunction{Length}{length}
\SetKwFunction{Max}{max}
\SetKwProg{Fn}{{\textbf{function}}}{}{}
\SetKw{KwRet}{{\textbf{return}}}

\DeclareRobustCommand{\origadd}[1]{#1}

\acrodef{ML}[ML]{machine learning}

\acrodef{DL}[DL]{deep learning}

\acrodef{NSF}[NSF]{National Science Foundation}

\acrodef{AI}[AI]{artificial intelligence}

\acrodef{AGI}[AGI]{artificial general intelligence}

\acrodef{FL}[FL]{Federated Learning}

\acrodef{CV}[CV]{computer vision}

\acrodef{CL}[CL]{Critical Learning}

\acrodef{CoV}[CoV]{Coefficient of Variation}

\acrodef{AC}[AC]{Attacking-Critical}

\acrodef{CAGR}[CAGR]{compound annual growth rate}

\acrodef{CCT}[CCT]{Center for Computation and Technology}

\acrodef{SLO}[SLO]{service level objective}

\acrodefplural{SLOs}[SLO]{service level objectives}

\acrodef{RL}[RL]{reinforcement learning}

\acrodef{DRL}[DRL]{deep reinforcement learning}

\acrodef{VM}[VM]{virtual machine}

\acrodefplural{VM}[VMs]{virtual machines}

\acrodef{microVM}[microVM]{micro virtual machine}

\acrodefplural{microVM}[microVMs]{micro virtual machines}

\acrodef{ITC}[ITC]{Innovation \& Technology Commercialization}

\acrodef{DAG}[DAG]{directed acyclic graph}

\acrodefplural{DAG}[DAGs]{directed acyclic graphs}

\acrodef{SFA}[SFA]{single point authentication}

\acrodef{HPC}[HPC]{high-performance computing}

\acrodef{SBIR}[SBIR]{Small Business Innovation Research}

\acrodef{IoT}[IoT]{Internet of Things}

\acrodef{DML}[DML]{distributed machine learning}

\acrodef{GNN}[GNN]{graph neural network}

\acrodefplural{GNN}[GNNs]{graph neural networks}

\acrodef{BSR}[BSR]{backdoor success rate}

\acrodef{BTA}[BTA]{backdoor task accuracy}

\acrodef{ATT}[ATT]{App Tracking Transparency}

\acrodef{DNN}[DNN]{deep neural network}

\acrodef{LLM}[LLM]{large language model}
\newcommand{\LLM}{\ac{LLM}\xspace}

\acrodefplural{LLM}{large language models}

\acrodef{DP}[DP]{Differential Privacy}

\acrodef{CNN}[CNN]{Convolutional Neural Network}

\acrodef{FLOP}[FLOP]{floating point operation}
\acrodefplural{FLOP}[FLOPs]{floating point operations}

\acrodef{MAC}[MAC]{multiply-and-accumulate}

\acrodef{FLOPS}[FLOPS]{Floating Point Operation per Second}

\acrodef{FC}[FC]{Fully Connected}

\acrodef{URM}[URM]{underrepresented minority}

\acrodef{HSSR}[HSSR]{High School Summer Research}

\acrodef{LSU}[LSU]{Louisiana State University}

\acrodef{LoRA}[LoRA]{Low-Rank Adaptation}
\newcommand{\LoRA}{\ac{LoRA}\xspace}

\acrodef{TTFT}[TTFT]{Time-To-First-Token}
\newcommand{\TTFT}{\ac{TTFT}\xspace}

\acrodef{TPOT}[TPOT]{Time-Per-Output-Token}
\newcommand{\TPOT}{\ac{TPOT}\xspace}

\acrodef{BO}[BO]{Bayesian optimization}

\acrodef{MAPE}[MAPE]{mean absolute percentage error}

\acrodef{CDF}[CDF]{cumulative distribution function}

\acrodef{GP}[GP]{Gaussian Process}

\acrodef{NLP}[NLP]{natural language processing}

\acrodef{VAE}[VAE]{variational auto-encoder}

\acrodefplural{VAE}[VAEs]{variational auto-encoders}

\acrodef{GAN}[GAN]{generative adversarial network}

\acrodefplural{GAN}[GANs]{generative adversarial networks}

\acrodef{KL}[KL]{Kullback-Leibler}

\acrodef{NAS}[NAS]{neural architecture search}

\acrodef{ILP}[ILP]{integer linear programming}

\acrodef{SM}[SM]{Streaming Multiprocessors}

\makeatletter
\renewcommand{\@maketitle}{\newpage
  \vbox{%
    \hsize=\textwidth
    \centering
    {\Large\bfseries \@title\par}%
    \vskip 8pt
    {\normalfont\normalsize \@author\par}%
    \vskip 10pt
  }%
}
\makeatother

\begin{document}

\title{ServerlessLoRA: Enabling Low-Latency Serverless Multi-LoRA Serving}
\author{%
  {\large
    \href{mailto:suiyifan@sjtu.edu.cn}{Yifan Sui}\textsuperscript{1}\quad
    \href{mailto:hwang9@stevens.edu}{Hao Wang}\textsuperscript{2}\quad
    \href{mailto:hyu42@stevens.edu}{Hanfei Yu}\textsuperscript{2}\quad
    \href{mailto:xkq@cse.ust.hk}{Kaiqiang Xu}\textsuperscript{3}\\[3pt]
    \href{mailto:yitao@tju.edu.cn}{Yitao Hu}\textsuperscript{4}\quad
    \href{mailto:chen-chen@sjtu.edu.cn}{Chen Chen}\textsuperscript{1}\quad
    \href{mailto:lijx@sjtu.edu.cn}{Jianxun Li}\textsuperscript{1}\quad
    \href{mailto:kaichen@cse.ust.hk}{Kai Chen}\textsuperscript{3}\endgraf}%
  \vspace{4pt}%
  {\normalsize\itshape
    \textsuperscript{1}Shanghai Jiao Tong University\quad
    \textsuperscript{2}Stevens Institute of Technology\\[2pt]
    \textsuperscript{3}iSING Lab, HKUST\quad
    \textsuperscript{4}Tianjin University\endgraf}%
}
\date{}
\maketitle
\begin{abstract}

Multi-LoRA (Low-Rank Adaptation) serving allows many specialized LLM variants to share the same base model by attaching lightweight adapters. This makes it attractive for serving large catalogs of domain-, tenant-, and task-specific models. However, existing deployments waste resources on rarely used adapters, while directly running LoRA inference on serverless platforms introduces high memory overhead, repeated cold starts, and poor scheduling decisions. 
This paper presents \sys, a serverless multi-LoRA serving system that separates serving state into shared backbone state, reusable variant warm state, and request-private state. \sys loads each backbone once, shares it read-only across isolated LoRA functions, \origadd{batches selected backbone operations,} selectively warms variant state, and routes requests based on state locality and GPU contention.
Evaluated on industrial traces, \sys reduces time-to-first-token by up to $92.3\%$ against serverless baselines and achieves $1.66$--$3.01\times$ and $2.20$--$3.26\times$ higher latency--cost efficiency than vLLM-LoRA and dLoRA, respectively.
\end{abstract}

\section{Introduction}
\label{sec:intro}

\LoRA has become a practical substrate for serving large numbers of specialized \LLM variants from a shared backbone~\cite{chen2024punica,sheng2023s,wu2024dlora}. Rather than deploying a separate model per domain, tenant, or task, providers increasingly maintain catalogs of lightweight adapters that attach to the same base model; this workflow is now supported across major cloud and AI serving stacks, including Azure AI Foundry~\cite{azure_ai_foundry_lora}, Google Vertex AI~\cite{google_vertex_ai_lora}, Amazon SageMaker~\cite{aws_sagemaker_lora}, NVIDIA NIM~\cite{nvidia_nim_lora}, Hugging Face PEFT~\cite{huggingface_peft_lora}, and vLLM~\cite{vllm_lora_adapters}. This design exposes a large asymmetry in the serving stack. A Llama3.1-8B backbone occupies roughly 16~GB in FP16, while a typical rank-8 adapter is only tens of MB~\cite{grattafiori2024llama}. Efficient multi-\LoRA serving therefore hinges on loading and amortizing the expensive backbone once, while switching only the small adapter state across requests.

\LoRA workloads are a natural match for serverless inference because adapter demand is both long-tailed and bursty. In a snapshot of 156{,}438 Hugging Face adapter repositories~\cite{huggingface_hub_api,peft}, the top 10\% account for 94.2\% of downloads, leaving most adapters accessed only sporadically. 
At the same time, production \LLM traces from Azure and ChatGPT show hourly demand spikes as high as $164\times$ the median rate~\cite{stojkovic2025dynamollm,wang2025burstgpt}. Conventional serverful serving\footnote{We use \emph{serverful} to refer to VM-based, long-running serving deployments such as vLLM~\cite{kwon2023efficient} and dLoRA~\cite{wu2024dlora}.} is poorly matched to this combination: 
it must keep backbones warm for the long tail and over-provision GPUs for bursts, wasting GPU resources and money. Serverless inference is built precisely for this workload: pay-per-use execution removes the cost of idle long-tail variants, scale-to-zero and fast scale-out absorb spikes within seconds, and function-level isolation gives flexible per-tenant management and accounting~\cite{amazon_bedrock,nvidia_dgx_platform,firecracker,fu2024serverlessllm}.

However, simply placing \LoRA inference behind a serverless interface does not deliver these benefits. The problem is a mismatch in granularity. Existing serverless runtimes manage state, scheduling, and isolation at the function-instance boundary~\cite{firecracker,fu2024serverlessllm,sui2024pre}. A \LoRA request, however, spans three distinct classes of state: a read-only \emph{backbone state} shared across variants, a reusable \emph{variant warm state} such as libraries, adapter weights, \origadd{and CUDA execution} state, and a \emph{request-private state} such as the key--value (KV) cache. 
Collapsing these states into a single function instance forces one scheduling and lifecycle policy onto state with different sharing, reuse, and isolation needs. This creates three sources of inefficiency (\S\ref{subsec:conflicts}):

\begin{itemize}[leftmargin=*,topsep=2pt,itemsep=2pt]
\item \textbf{Spatial inefficiency:} function-level isolation replicates the frozen backbone across containers and CUDA contexts, making GPU memory scale with active variants rather than base models.
\item \textbf{Temporal inefficiency:} scale-to-zero discards reusable variant warm state, including libraries, adapters, \origadd{and CUDA execution state}. Reconstructing this state is costly and can dominate \TTFT for long-tail adapters.
\item \textbf{Scheduling inefficiency:} instance-level scheduling ignores backbone locality, warm-state locality, and high-bandwidth memory (HBM) bandwidth contention, causing redundant weight movement, interference, and higher latency under bursts.
\end{itemize}

However, none of the existing solutions can address all three inefficiencies simultaneously, because state management remains coupled to coarse execution boundaries rather than the natural granularity of \LoRA serving state.
Serverful multi-\LoRA systems~\cite{chen2024punica,sheng2023s,wu2024dlora} share backbones within long-lived serving processes, while serverless \LLM systems~\cite{fu2024serverlessllm,sui2024pre,lou2026hydraserve,yu2026faascale} reduce cold-start overheads for isolated functions.
More fundamentally, neither line of work jointly accounts for state reuse and HBM-bandwidth contention when scheduling isolated \LoRA functions (\S\ref{subsec:limitations}).

In this paper, we present \sys, a serverless multi-LoRA serving system that can address all three inefficiencies simultaneously by decoupling state management from the function-instance boundary. \sys is based on the observation that a \LoRA request touches state with distinct scopes and lifetimes: the backbone is large, read-only, and shared across variants; variant warm state is smaller but expensive to reconstruct and should be reused when beneficial; request-private state, such as KV cache, must remain isolated. 
\sys manages each state class at its natural granularity through three coordinated mechanisms:

\begin{itemize}[leftmargin=*,topsep=2pt,itemsep=2pt]
\item \textbf{Backbone sharing} (\S\ref{subsec:backbone_sharing}) \origadd{uses a \emph{Backbone Function} to combine read-only weight sharing with batched backbone execution across isolated \LoRA functions. It} loads each base model once\origadd{, exports its weights} read-only\origadd{, and batches selected backbone operations through a platform-trusted executor while preserving function isolation}.
\item \textbf{Artifact warming} (\S\ref{subsec:warming}) reuses variant warm state by \origadd{preparing and retaining} libraries, adapter weights, \origadd{and CUDA execution state} on existing idle containers and GPUs, rather than maintaining always-warm replicas. An \emph{Artifact-Warming Scheduler} decides which warm states to retain and where to place them under \origadd{long-tailed adapter demand}.
\item \textbf{Locality- and contention-aware scheduling} (\S\ref{subsec:scheduling_design}) \origadd{coordinates placement, local batching, and batched backbone execution using backbone locality, variant} warm-state \origadd{readiness}, and GPU contention. \origadd{During bursts, it reclaims idle GPU state while retaining reusable preparation where possible}.
\end{itemize}

We implement \sys as a serverless multi-\LoRA serving runtime. 
To capture realistic invocation dynamics, we evaluate \sys using production Azure Functions traces that span predictable, normal, and bursty arrival patterns. 
We compare with serverless systems InstaInfer~\cite{sui2024pre}, ServerlessLLM~\cite{fu2024serverlessllm}, and HydraServe~\cite{lou2026hydraserve}, a naive CUDA interprocess communication (IPC) backbone-sharing baseline, and serverful systems dLoRA~\cite{wu2024dlora} and vLLM~\cite{kwon2023efficient} in merged-model and multi-LoRA configurations.

This paper makes the following contributions:

\begin{itemize}[leftmargin=*,topsep=2pt,itemsep=2pt]
\item We show that naive serverless multi-\LoRA serving is inefficient because it binds heterogeneous serving state to the function-instance boundary, causing backbone duplication, repeated reconstruction of variant warm state, and state-oblivious scheduling.

\item We design \sys, a serverless multi-\LoRA serving system that separates backbone state, variant warm state, and request-private state using \origadd{Backbone Functions for read-only weight sharing and batched backbone execution}, flexible artifact warming, and locality- and contention-aware scheduling with burst-time offloading.

\item We implement \sys and evaluate it on industrial traces, showing up to 92.3\% lower \TTFT against serverless baselines and 6.25--9.38$\times$ the cost-effectiveness of vLLM merged.
\end{itemize}

\begin{figure}[t]
\centering
\includegraphics[width=\linewidth]{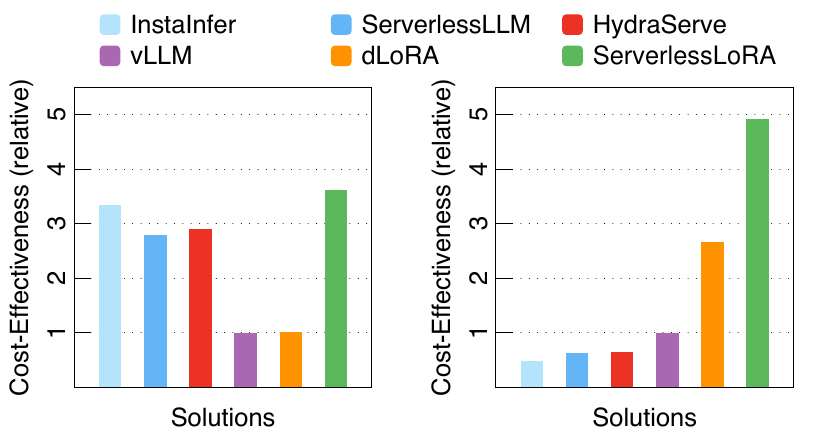}
\par\nointerlineskip
\captionsetup{skip=2pt}
\captionsetup[subfigure]{skip=0pt}
\begin{subfigure}[t]{0.49\linewidth}
    \centering
    \caption{Cost-effectiveness of serverless and serverful solutions for 1 Llama2-7B base LLM.}
    \label{fig:moti1}
\end{subfigure}
\hfill
\begin{subfigure}[t]{0.49\linewidth}
    \centering
    \caption{Cost-effectiveness of serverless and serverful solutions for 4 Llama2-7B LoRA LLMs.}
    \label{fig:moti2}
\end{subfigure}

\caption{Cost-effectiveness of serverless and serverful solutions (normalized to vLLM).}
\label{fig:motiv}

\vspace{4pt}
\end{figure}

\section{Background and Motivation}
\label{sec: background_v3}

Serverless is an attractive paradigm for multi-\LoRA inference, but naively combining them raises challenges that neither existing serverless nor serverful systems can solve.

\subsection{Low-Rank Adaptation (LoRA)}
\label{subsec:lora}

\LoRA is a parameter-efficient fine-tuning method that freezes a large backbone \LLM and learns small low-rank updates for selected weight matrices. Concretely, for a weight matrix \(W \in \mathbb{R}^{h \times d}\), LoRA learns \(A \in \mathbb{R}^{h \times r}\) and \(B \in \mathbb{R}^{r \times d}\) with \(r \ll \min(h,d)\), yielding a separable forward pass:
\[
x(W+AB) = xW + xAB.
\]
For example, a Llama3.1-8B backbone in FP16 occupies about 16 GB GPU memory, while a typical rank-8 LoRA adapter is only tens of MB~\cite{grattafiori2024llama}.

Serving a variant also produces reusable per-variant runtime state---adapter weights placed on the GPU, adapter-to-backbone metadata, and CUDA \origadd{execution state, including kernels and reusable CUDA Graphs}. Although much smaller than loading the backbone, this setup is not free and can be reused across later requests to the same variant. We call it the \emph{variant warm state}, distinct from both the shared backbone and per-request state such as KV cache.

\subsection{Serverless for LLM Inference}
\label{subsec:serverless_llm}

Serverless computing offers two key advantages over traditional \emph{serverful} deployments \origadd{for LLM inference}:

\textbf{(1) Better cost and GPU efficiency.} Serverless bills per use, avoiding idle GPU reservation under low request rates, which matches the long-tailed LoRA access pattern~\cite{huggingface_hub_api,peft}. We use cost-effectiveness\footnote{Following prior work~\cite{thorpe2021dorylus,wang2020infinicache,jia2021nightcore}, cost-effectiveness is defined as \(1/(\text{E2E\_latency} \times \text{Cost})\).} as a joint metric of end-to-end (E2E) latency and cost. Our evaluation (Fig.~\ref{fig:moti1}) shows that for Llama2-7B, serverless solutions~\cite{fu2024serverlessllm,sui2024pre} are \(3\times\) more cost-effective than representative serverful systems~\cite{kwon2023efficient,wu2024dlora}.

\textbf{(2) Superior elasticity under bursty load.} Serverless platforms scale out within seconds, while serverful systems operate on coarser minute-level control loops, making them slower to absorb bursts without over-provisioning.

\subsection{The Serverless--LoRA Serving Mismatch}
\label{subsec:conflicts}
\textbf{Root Cause.} Today's serverless runtimes use the function instance as the single unit of loading, residency, reclamation, scheduling, and isolation. This is too coarse for multi-adapter LoRA serving, where \origadd{shared \emph{backbone state},} reusable \emph{variant warm state} \origadd{(\S\ref{subsec:lora}), and \emph{request-private state} have different sharing, reuse, and isolation needs}. Forcing all three into the function-instance boundary creates three sources of inefficiency.

\noindent\textbf{Spatial inefficiency: Isolation prevents backbone sharing.}
Serverless executions typically run in isolated containers or microVMs~\cite{firecracker}, each with its own state and CUDA context, to enable security, fault containment, and accurate accounting. In contrast, efficient LoRA serving assumes that many adapters can reuse a single identical, read-only backbone in GPU memory. When isolation prohibits such sharing, each function instance must load its own copy of the backbone, causing GPU memory consumption and cost to scale roughly linearly with the number of concurrently served LoRA variants---even though the backbone parameters dominate the model size and are shared across adapters. For example, four Llama2-7B LoRA functions consume about 56.2~GB of GPU memory (four 14~GB FP16 backbones plus four 50~MB rank-8 adapters) instead of the 14.2~GB needed if the backbone were shared, a $4\times$ duplication. As Fig.~\ref{fig:moti2} shows, if each function loads an entire backbone, the cost-effectiveness of serverless is much lower than that of serverful solutions due to memory duplication.

\begin{figure}[t]
        \centering
        \includegraphics[width=\columnwidth,trim={5.5 3.1 10.3 3.1},clip]{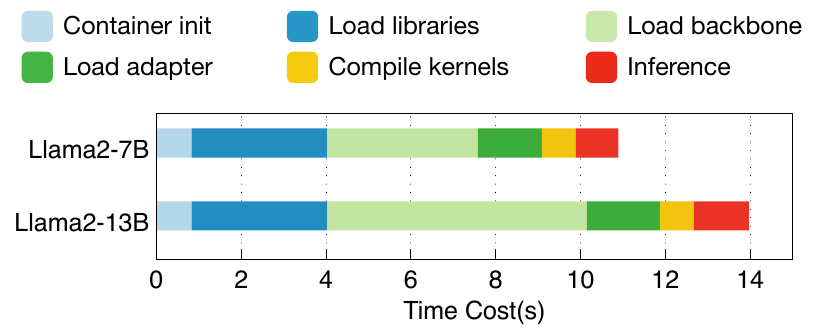}
        
        \caption{Time breakdown of a \LoRA \LLM inference with serverless computing.}
        \label{fig:motivation_coldstart_breakdown}
        
\vspace{4pt}
\end{figure}

\begin{figure}[t]
\centering
\includegraphics[width=\linewidth,trim={2.4 6.8 0 5.6},clip]{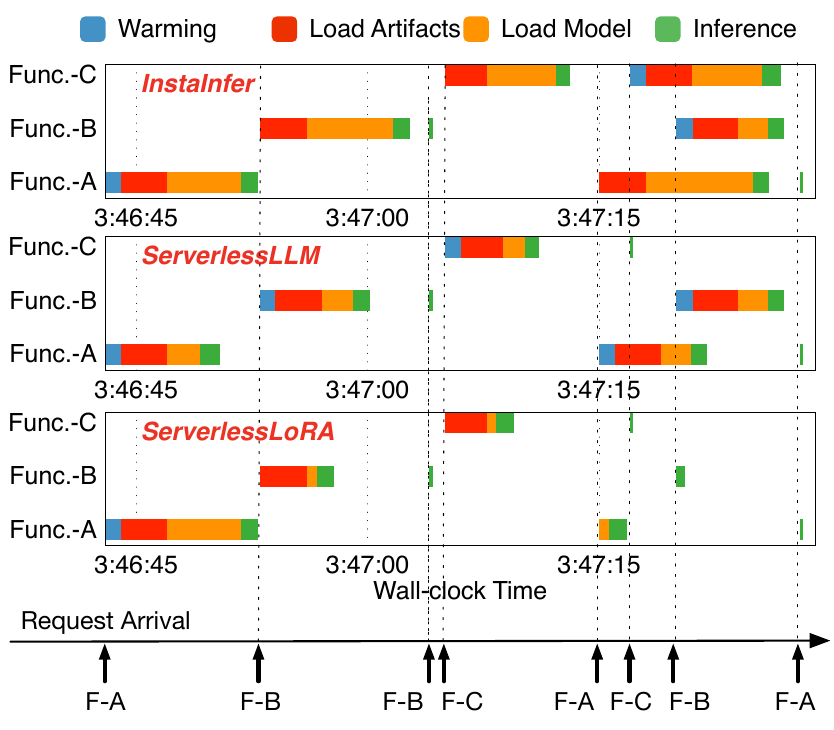}

\caption{LoRA function invocations' time breakdown.}
\label{fig:timeline}

\vspace{4pt}
\end{figure}

\noindent\textbf{Temporal inefficiency: Statelessness forces repeated cold starts.}
Serverless environments utilize short-lived stateless instances and aggressive scale-to-zero, whereas LoRA inference requires a ``hot'' state to keep \TTFT low. Under bursty workloads, repeated cold starts force rarely accessed variants in the long-tailed access pattern to lose and rewarm their variant warm state, negating the latency advantages of lightweight adapters. Crucially, even if we share the backbone model to skip the backbone loading phase entirely, reconstructing the remaining variant warm state---framework libraries, CUDA runtime, and kernel compilation---still incurs substantial cold-start latency that significantly increases \TTFT. As Fig.~\ref{fig:motivation_coldstart_breakdown} shows, loading the 7B backbone takes 3.8~seconds, but initializing these non-model artifacts takes a dominating 6~seconds; in total, non-inference overhead accounts for over 90\% of cold invocation time. As a concrete example, Fig.~\ref{fig:timeline} visualizes the E2E latency breakdown of three Llama2-13B LoRA functions under the Azure Trace: both InstaInfer and ServerlessLLM exhibit high cold-start latency due to repeated backbone and LoRA loading.

\noindent\textbf{Scheduling inefficiency: State-oblivious routing and execution.}
Traditional serverless platforms fail multi-adapter LoRA serving because their schedulers choose among function instances without enough information about the model state and resources behind them.

\noindent\textit{Locality-blind routing:} Load balancers route based on container warmth and resource capacity, thereby sending requests to GPUs lacking the required backbone or variant warm state and triggering heavy cold-starts.

\noindent \textit{Contention-blind execution:} \origadd{Existing runtimes execute isolated LoRA functions independently, without coordinating backbone operations across function boundaries}. When scheduled on the same GPU, \origadd{these functions} redundantly fetch the same backbone weights from HBM to compute units at the same time. \origadd{Small per-function batches from long-tail variants further limit weight reuse.
With backbone data sharing alone, increasing the number of colocated Qwen3-8B LoRA functions from 1 to 4 raises median time-per-output-token (TPOT)\acused{TPOT} from 18.21 to 72.84 ms (4.00$\times$)}.

\subsection{Limitations of Existing Solutions}
\label{subsec:limitations}
\noindent\textbf{Serverful multi-LoRA systems.} Solutions like Punica~\cite{chen2024punica}, S-LoRA~\cite{sheng2023s}, and dLoRA~\cite{wu2024dlora} address efficiency by colocating many variants around a single long-lived backbone. This improves sharing and batching, but \origadd{places tenant-specific serving state within a common serving process rather than independently isolated function instances}. It also assumes warm GPU residency for long periods, which conflicts with rarely accessed, long-tailed variants.

\noindent\textbf{General serverless LLM systems.} Existing serverless inference systems primarily optimize artifact loading and cold-start latency~\cite{fu2024serverlessllm,sui2024pre}. Each function instance, however, still owns the model state it serves, so they fail to address cross-function redundancy in multi-adapter deployments where many LoRA variants share the same backbone.

\section{\sys: A Bird's-Eye View}
\label{sec:sys_overview_v2}

\sys is a serverless \origadd{multi-LoRA} inference system that resolves the three sources of inefficiency in \S\ref{sec: background_v3} by managing each kind of serving state separately\origadd{.} The frozen backbone is identical across variants \origadd{and can be} shared read-only while keeping function-specific state isolated. Serverless platforms keep ``keep-alive'' instances post-invocation, \origadd{providing idle capacity for opportunistically warming} variant warm state~\origadd{\cite{gunasekaran2020fifer,zhang2021faster,yu2023libra,zhou2022aquatope,romero2021infaas,shahrad2020serverless}.} LoRA's mathematical separability (\S\ref{subsec:lora}) permits unmerged inference, allowing \origadd{selected backbone operations to be batched across functions while adapter computation remains function-local}~\cite{chen2024punica,wu2024dlora,sheng2023s}.

\subsection{\sys's System Architecture}
\label{subsec:architecture}

\begin{figure}[t]
\centering
\includegraphics[width=\columnwidth]{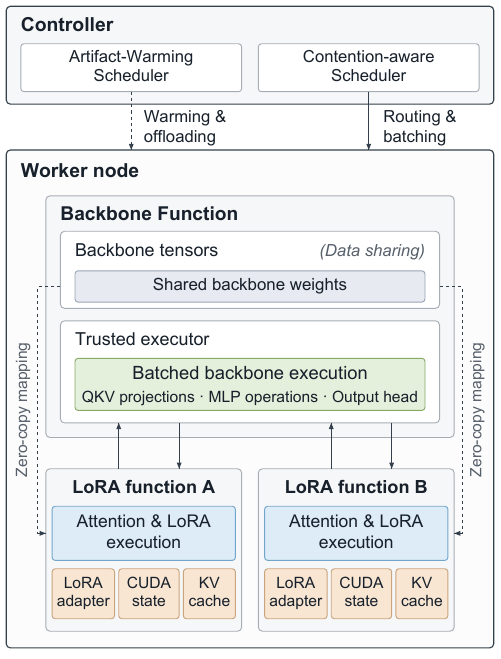}
\captionsetup{labelfont=bf}
\caption{Overview of \sys's architecture.}
\label{fig:system_design}
\vspace{4pt}
\end{figure}

\begin{figure*}[t]
\centering
\includegraphics[width=\textwidth]{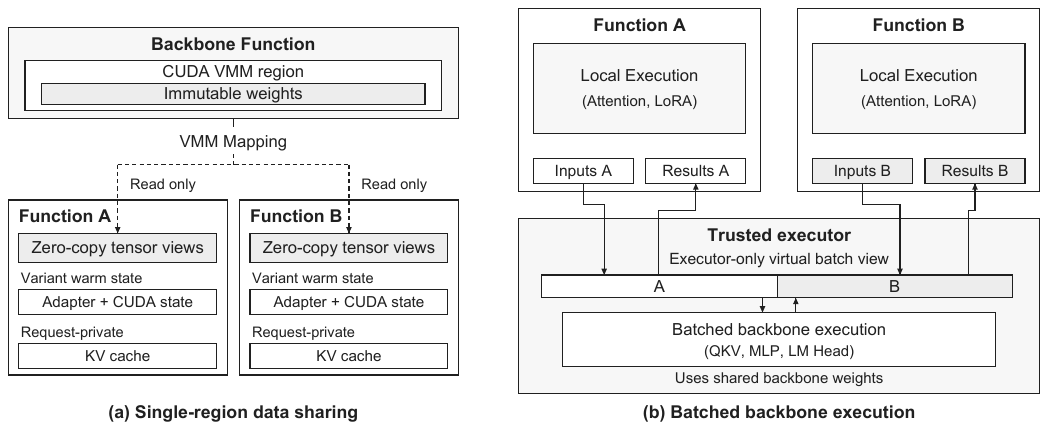}
\caption{Backbone sharing through (a) single-region weight mapping and (b) batched backbone execution.}
\label{fig:backbone_sharing_v3}
\vspace{4pt}
\end{figure*}

\origadd{Figure~\ref{fig:system_design} shows \sys's system architecture. Four components manage serving state and coordinate execution.}

\noindent\origadd{\textbf{Backbone Functions.} To address spatial inefficiency, Backbone Functions eliminate redundant backbone residency. They own a single copy of each backbone per GPU,} export it as read-only shared tensors to isolated LoRA functions\origadd{, and support batched backbone execution through a platform-trusted executor}.

\noindent\textbf{Artifact-Warming Scheduler.} \origadd{To address temporal inefficiency, the Artifact-Warming Scheduler repurposes idle capacity to minimize \TTFT under scale-to-zero. Running in the controller, it} makes holistic pre-warming decisions \origadd{for shared backbone state and reusable variant warm state} under resource constraints\origadd{. Worker-side Artifact-Warming Agents execute these decisions} on existing idle containers/GPUs.

\noindent\textbf{\origadd{Burstiness-aware Offloader}.} \origadd{To address scheduling inefficiency under bursts, the Burstiness-aware Offloader reclaims GPU memory from idle functions for active requests. Running on each worker, it releases idle KV memory or retires GPU execution state while retaining useful CPU preparation}.

\noindent\textbf{\origadd{Contention-aware Scheduler}.} \origadd{The Contention-aware Scheduler addresses scheduling inefficiency caused by uncoordinated execution among functions sharing a backbone. Running in the controller, it coordinates backbone-aware placement, per-function batching, and batched backbone execution to reduce per-function latency (\S\ref{subsec:scheduling_design})}.

\subsection{System Workflow}
\label{subsec:workflow}

\noindent\textbf{Backbone Sharing\origadd{.}} \origadd{\sys creates a Backbone Function on demand to load a single copy of the backbone. Colocated LoRA functions attach to it through} read-only \origadd{zero-copy mappings, avoiding separate backbone copies while keeping function-specific state isolated.}

\noindent\origadd{\textbf{Artifact Warming.} Before request arrival, the} Artifact-Warming Scheduler \origadd{selects warming targets based on} resource availability and function frequency\origadd{. Worker-side Artifact-Warming Agents prepare and retain selected backbone weights, libraries, adapter data, and CUDA execution state on existing idle resources. Under GPU memory pressure, the worker-side Burstiness-aware Offloader reclaims idle GPU state and feeds updated warm-state information back to the scheduler.}

\noindent\textbf{Contention-aware Routing and Batching\origadd{.}} On request arrival, the Contention-aware Scheduler picks a target based on backbone residency, warm-state availability, free GPU memory, and contention. Requests are then dispatched to an isolated \origadd{LoRA function. During inference, LoRA functions form local batches and execute attention and low-rank adapter projections locally. The trusted executor within the Backbone Function combines compatible local batches for batched backbone execution and returns results to each function, reducing redundant weight reads across functions.}

\section{\origadd{\sys's Design}}
\label{sec:sys_design}

This section presents three coordinated mechanisms that address the sources of inefficiency identified in \S\ref{sec: background_v3}. Backbone sharing separates shared backbone state and selected backbone operations from function-local execution (\S\ref{subsec:backbone_sharing}). Artifact warming preserves useful preparation across invocations (\S\ref{subsec:warming}). State-aware runtime coordination uses these distinctions to jointly manage locality, contention, and memory pressure (\S\ref{subsec:scheduling_design}).

\subsection{Backbone Sharing: State and Execution}
\label{subsec:backbone_sharing}

To resolve spatial inefficiency, \sys introduces the Backbone Function: a per-backbone serverless role that owns the base model as a single immutable GPU-resident object per GPU. It combines read-only weight sharing with batched backbone execution while preserving isolation of function-specific state (Figure~\ref{fig:backbone_sharing_v3}).

The Backbone Function is created on demand and kept warm according to expected reuse and available idle capacity (\S\ref{subsec:warming}). Runtime coordination adjusts the number and placement of Backbone Function instances to the working set and queue pressure, including scale-to-zero when idle (\S\ref{subsec:scheduling_design}). Before releasing a backbone replica's physical weights, \sys drains accepted work and closes all dependent LoRA functions, discarding their retained GPU execution state and CPU preparation.

\paragraph{Sharing principle.} \sys shares immutable backbone weights and batches backbone query/key/value (QKV) projections, multilayer perceptron (MLP) operations, and the output head across LoRA functions, while keeping attention and low-rank adapter computation function-local. Sharing weights avoids per-function backbone copies, but residency sharing alone still leaves functions fetching the same weights for separate matrix multiplications. The resulting HBM-bandwidth contention slows individual functions and reduces aggregate throughput, motivating batched backbone execution.

The boundary of batched backbone execution accounts for both request-private state and cross-function synchronization. Attention accesses and updates the KV cache, whereas backbone QKV projections, MLP operations, and the output head operate on submitted activations. \sys therefore batches these operations while keeping QKV LoRA updates, positional processing, attention, KV updates, and the attention output projection function-local. This boundary reduces redundant weight reads while limiting cross-function synchronization. Function-local backbone operations still use the shared read-only weights.

\paragraph{Single-region data sharing.} A natural approach is to export each backbone tensor through CUDA IPC, as in prior serverless data-sharing systems~\cite{li2022tetris,cai2023smss,gu2023fast,cui2025efficient}. For a large backbone, this incurs thousands of cross-context imports, increasing cold-start latency and GPU memory overhead. The Backbone Function instead allocates one contiguous region using CUDA virtual memory management (VMM) and loads the checkpoint's weights at fixed offsets. It publishes a single IPC handle and a layout manifest describing the tensors. Each LoRA function imports the region once, maps it read-only, and reconstructs tensor views locally. This avoids per-function weight copies and replaces per-tensor imports with one region import and local metadata resolution.

\paragraph{Reducing redundant weight reads through cross-function batching.} Each LoRA function delegates backbone QKV projections, MLP execution, and the output head to a platform-trusted executor during both prefill and decode. Compatible functions contribute their entire local batches, forming a combined batch that reduces redundant weight reads across functions. Each function supplies its input activations and the private LoRA increments needed by the delegated operations. The executor follows the backbone's original operators and dependencies and incorporates these increments at their original application points. It returns the outputs and intermediate activations needed to complete private computation. LoRA weights and low-rank projections remain local. For example, in a SwiGLU MLP, each function computes Gate/Up LoRA increments from the MLP input before delegation, and the executor incorporates them before SwiGLU. Each function then uses the returned SwiGLU activation to compute the Down LoRA increment and adds it locally to the returned backbone Down output before the layer's residual path, without another delegation. The policy for admitting compatible batches is described in \S\ref{subsec:scheduling_design}.

To reduce host-side submission overhead, each function replays its complete per-iteration execution path, including private computation and cross-function synchronization, as a private CUDA Graph within its own CUDA context. A separate executor graph captures the batched backbone operations. Stream-ordered signals connect these graphs, removing host-mediated handoffs between layers. Replaying prepared graphs avoids repeated per-operator dispatch and kernel-launch setup on the CPU.

A cross-function iteration binds a set of functions and their input, output, and control mappings. Compatible bindings are reused across successive iterations. Changing the set of bound functions requires a new binding, but does not require rebuilding a compatible executor template (\S\ref{subsec:warming}). Separately prepared local graphs support execution without cross-function batching.

\paragraph{Cross-function data exchange.} Batched backbone execution must exchange activations across function boundaries without exposing one function's data to another. Each function has physically separate input and output buffers with associated control mappings. Only the executor maps the bound functions' submitted rows into a combined batch view in its own virtual address space, forming a virtual matrix with a uniform row stride. Batched backbone GEMMs directly consume the corresponding input and intermediate views, with the batch dimension spanning all submitted rows. This avoids a gather copy into a common input buffer, while publishing inputs and consuming results still incur function-local data movement. A function can write its own input and read its own output, but cannot obtain another function's mappings. Stream-ordered ready/done signals connect private input production, batched backbone execution, and result consumption. Section~\ref{subsec:security} describes the enforcement of this boundary.

Rebinding and reclamation wait for the executor and all bound functions to finish their dependent work, including final result consumption.

\paragraph{Multi-GPU backbones.} For backbones spanning multiple GPUs, \sys applies the same sharing boundary to each tensor- or pipeline-parallel partition. Each GPU's backbone weights are shared read-only across functions, while LoRA weights and KV caches remain function-private under the corresponding parallel layout. Batched backbone execution maintains the same combined batch and request order across tensor-parallel ranks. This preserves vLLM's tensor-parallel execution structure, allowing \sys to reuse its parallel operators and collective communication for efficient multi-GPU execution.

\subsection{Artifact Warming}
\label{subsec:warming}

Artifact warming prepares and retains shared backbone state and reusable variant warm state, covering both data loading (libraries, backbones, and adapters) and CUDA execution-state preparation.

\paragraph{Warming libraries and models.} \sys preloads libraries in idle containers and retains backbone models and LoRA adapters in CPU or GPU memory. Backbone Functions participate in warming under the same policy as other artifacts: their GPU-resident models are retained according to expected reuse and shared across colocated LoRA functions. The scheduler places each adapter's GPU warm state on the same GPU as its corresponding Backbone Function, so later invocations can reuse both rather than reload the backbone elsewhere.

\paragraph{Warming CUDA execution state.} Resident libraries and weights do not by themselves make a LoRA function ready to execute. \sys prepares CUDA runtime state, kernels, and private graphs for local execution or batched backbone execution, together with reusable executor templates. These templates capture the executor's batched backbone operations and depend on the backbone, number of bound functions, and batch capacity rather than function or request identities. Different function combinations can therefore reuse a prepared template through function-specific bindings.

A bounded set of capacity shapes covers neighboring batch sizes, trading limited additional computation for fewer templates and lower preparation costs. The Artifact-Warming Scheduler uses historical demand to choose shapes worth preparing and retaining, and tracks readiness for private graphs and batched backbone execution separately. Private graphs require valid KV mappings, while batched backbone execution also requires a ready template and function-specific preparation and binding.

\paragraph{Joint warming policy.} To avoid extra provisioning costs, \sys warms state only on existing idle resources. The scheduler allocates this capacity across both kinds of preparation using a common value-density policy. Each artifact or reusable execution state has a memory requirement and a pre-warming benefit, estimated as the loading or initialization delay avoided multiplied by the predicted probability of a function invocation that can reuse the state. The delay is estimated from offline profiling and historical logs. Following prior work~\cite{sui2024pre}, we model invocations as a Poisson process with arrival rate \(\lambda_f\) estimated from recent invocation history. The probability of at least one invocation within a look-ahead window \(T\) is \(p_f=1-e^{-\lambda_f T}\). Selection is formulated as a Precedence-Constrained Knapsack Problem (PCKP). Capacity constraints bound container and GPU memory; loading-order constraints require libraries before models and models before CUDA kernels; backbone-adapter coupling requires GPU-resident adapters to be colocated with their corresponding Backbone Function. Execution-state preparation additionally respects the template, shape, and function-specific dependencies described above. Shared allocations are counted once across dependent functions. A greedy heuristic selects feasible artifact-location pairs by value density \(\rho_i^f = v_i^f / w_i^f\), where \(v_i^f\) is the expected pre-warming benefit and \(w_i^f\) is the memory footprint. This favors preparation that avoids more cold-start delay per unit memory while respecting the prerequisites that make it usable.

\paragraph{Overlapping preparation.} Once warming targets are selected, \sys uses separate channels for CPU preparation and GPU maintenance. Library preparation for one function can overlap GPU activation and graph preparation for another, while already-ready functions continue serving.

\subsection{State-aware Runtime Coordination}
\label{subsec:scheduling_design}

\sys coordinates placement, batching, and offloading to reuse backbone and variant warm state while managing GPU contention and memory pressure.

\paragraph{Backbone-aware placement.} \sys colocates each LoRA function with a matching Backbone Function, balancing state reuse against GPU contention. Among GPUs satisfying memory and waiting limits, denoted by \(\mathcal{G}_f\), it selects

\[
g^* = \arg\min_{g\in\mathcal{G}_f}
\left(\lVert\mathbf{u}_g+\Delta\mathbf{u}_{f,g}\rVert_2^2-\lVert\mathbf{u}_g\rVert_2^2\right),
\]

where \(\mathbf{u}_g\) contains normalized utilization measures for memory, service capacity, and function concurrency, and \(\Delta\mathbf{u}_{f,g}\) is the increase from placing function \(f\). Memory estimates use profiling and current residency; service-load estimates use recent arrival rates, queue lengths, and same-backbone completion throughput, with profiling for initialization. Ties favor warm-state reuse.

Estimated waiting adds profiled remaining preparation time to the larger of the function-level and GPU-level queueing estimates. Batch execution times use an exponentially weighted moving average, initialized from profiling.

\sys activates additional LoRA instances and, when necessary, matching Backbone Functions on free GPUs when the working set exceeds current capacity. Queue pressure also triggers scale-out when the oldest request reaches a configured waiting limit and an additional replica is expected to shorten queue-draining time after accounting for its loading delay. When no GPU is free, drained instances may be replaced according to warm-state value.

\begin{figure}[t]
\centering
\includegraphics[width=\linewidth]{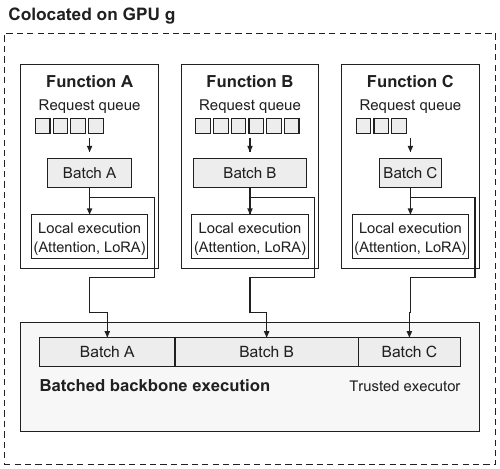}
\caption{Combining function-local batches for batched backbone execution.}
\label{fig:local_shared_batching}
\vspace{4pt}
\end{figure}

\paragraph{Contention-aware batching.} \sys coordinates function-local batching with batched backbone execution (Figure~\ref{fig:local_shared_batching}). Each function uses fill-or-expire admission: a batch becomes eligible at the batch limit or when a fixed timer from the oldest request's arrival expires. On GPU \(g\), function \(f\) selects the largest feasible batch from the head of its queue:

\[
b_f^* = \max\left\{b\in\mathcal{B}_f:
\Delta M_f^{\mathrm{KV}}(b)+\Delta M_f^{\mathrm{exec}}(b)\leq M_g^{\mathrm{free}}\right\},
\]

where \(\mathcal{B}_f\) contains batch sizes allowed by the queue length, execution shapes, and batch limit. The memory terms cover additional KV backing and execution-state preparation plus scratch space, respectively; \(M_g^{\mathrm{free}}\) is available GPU memory.

Among batches ready for batched backbone execution (\S\ref{subsec:warming}), \sys combines whole local batches using the same backbone, favoring earlier-ready work subject to limits on bound functions and total rows. Functions advertise their next eligible iteration during current execution; the bound functions remain fixed within an iteration and may change at the next boundary. Unselected batches execute locally when their local graphs are ready, including during binding preparation, without a fixed partner-waiting interval. Both paths obey memory and waiting limits.

\begin{figure}[t]
\centering
\includegraphics[width=\linewidth]{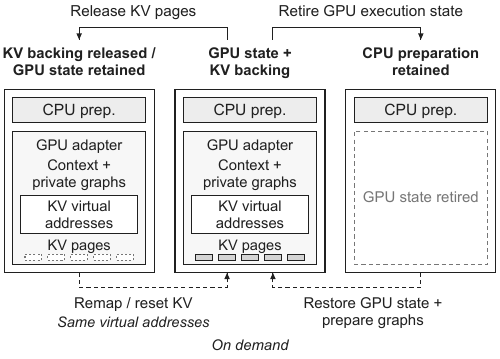}
\caption{GPU-state reclamation and restoration; dashed arrows indicate on-demand restoration.}
\label{fig:state_retention}
\vspace{4pt}
\end{figure}

\paragraph{Burstiness-aware offloading.} When the estimated incremental peak GPU memory required to prepare and execute incoming work exceeds available capacity, \sys reclaims idle state through two actions (Figure~\ref{fig:state_retention}): releasing KV backing while preserving its virtual addresses and reusable GPU execution state, or retiring GPU execution state while retaining useful CPU preparation.

For an eligible action removing state \(i\) from function \(f\), the offloader computes \(\rho_i^f = v_i^f / w_i^f\), where \(w_i^f\) is the GPU memory released and \(v_i^f\) is the recent request rate multiplied by the profiled restoration delay, including execution-state preparation. It selects lower-density actions until the memory deficit is covered, counting released allocations only once. Recently reclaimed GPU state is not warmed again without actual demand. Reclamation updates readiness information for subsequent placement and batching.

\subsection{Security and Isolation}
\label{subsec:security}

\paragraph{Trust domains and GPU assignment.} \sys limits GPU sharing to a tenant trust domain: LoRA functions permitted by the tenant's security policy to share GPUs. The provider and its managed serving runtime are trusted; tenant applications run in sandboxes without direct GPU access. During each allocation epoch, each physical GPU serves one trust domain exclusively. Backbone mappings, executor queues, and cross-function batches are domain-scoped. Functions retain separate processes, CUDA contexts, and private serving state.

\paragraph{Protecting the shared backbone.} Only functions legitimately scheduled by the platform can attach to a Backbone Function, enforced through OS-level peer authentication. Hardware-enforced read-only mappings allow functions to use the common weights without modifying the model used by other functions.

\paragraph{Isolating batched backbone execution.} Adapter weights, KV caches, attention state, and private graph pools remain inside separate function processes and CUDA contexts. The trusted executor accesses read-only backbone weights and explicitly authorized activation and control buffers, including submitted LoRA increments. Directional permissions restrict each function to its own input, output, and control rows. The executor accepts fixed, model-defined operations rather than tenant-supplied code. It validates submission metadata and confines each function's inputs, increments, and outputs to that function's batch rows. Batched backbone execution therefore does not expose another function's private buffers through the sharing interface. This access-control boundary is distinct from timing isolation.

\paragraph{Preserving isolation across reuse.} Buffer reuse must not turn an old handle into access to a new tenant's state. Reused buffer slots receive fresh, zeroed physical allocations rather than merely a new identifier. Old outputs and mappings are reclaimed only after dependent private and executor work completes and no retained graph binding references those mappings (\S\ref{subsec:backbone_sharing}). Stale handles continue to refer only to the retired physical generation. This preserves the same isolation boundary when functions depart and their execution resources are reused.

\section{Implementation}
\label{sec:implement}

We implement \sys in Python and C++/CUDA. To enable secure backbone sharing while preserving function isolation (\S\ref{subsec:security}), we develop a customized runtime for shared weight access, cross-function batching, and function lifecycle management. Its decoding backend reuses optimized FlashAttention kernels provided by vLLM~\cite{kwon2023efficient}. PyTorch and Hugging Face Transformers provide model definitions, weight loading, and basic tensor operations. Function-local LoRA computation, normalization, and KV updates use custom CUDA kernels. A Python controller coordinates request placement, artifact warming, memory reclamation, and backbone scaling.

\paragraph{Backbone weight sharing.} We integrate CUDA VMM with PyTorch's parameter loader through a C++ bridge. Each Backbone Function stores its weights in a single exportable GPU allocation and distributes the handle over a Unix-domain socket with peer-credential checks. LoRA workers import the allocation once, map it read-only, and reconstruct tensor views from a manifest of tensor offsets, shapes, and data types.

\paragraph{Batched backbone execution.} Model-layer hooks delegate backbone QKV projections, MLP operations, and the LM head to a trusted C++/CUDA executor, while attention and low-rank adapter computation remain function-local. The executor uses cuBLAS for batched projections and accesses per-function GPU mailboxes through a combined virtual batch view, avoiding a gather copy into a centralized input buffer. Both prefill and decode use CUDA Graphs: each worker captures its per-iteration execution path in a private graph, while a separate executor graph captures the batched backbone operations. CUDA stream-ordered wait/write operations coordinate input production, shared execution, and result consumption without per-layer host handoffs.

\paragraph{Artifact warming and reclamation.} A CUDA-uninitialized parent process retains imported libraries, the tokenizer, and CPU adapter weights, while a GPU child owns the function's CUDA context and execution state. Private KV storage uses VMM, allowing idle workers to release physical pages while preserving the virtual addresses referenced by retained graphs. Under greater memory pressure, the runtime terminates the GPU child while retaining CPU preparation, or evicts the entire function. Separate CPU-preparation and GPU-maintenance queues allow these operations to overlap with inference in ready functions.

\section{Evaluation}
\label{sec: eval}

\subsection{Experimental Setup}

\noindent\textbf{Testbed.} We use a 4-node cluster with 384 CPU cores, 2,048~GB memory, and 32 NVIDIA A100-80G GPUs.

\begin{figure}[t]
\centering
\includegraphics[width=\linewidth,trim={2.3 2.6 13.1 7.3},clip]{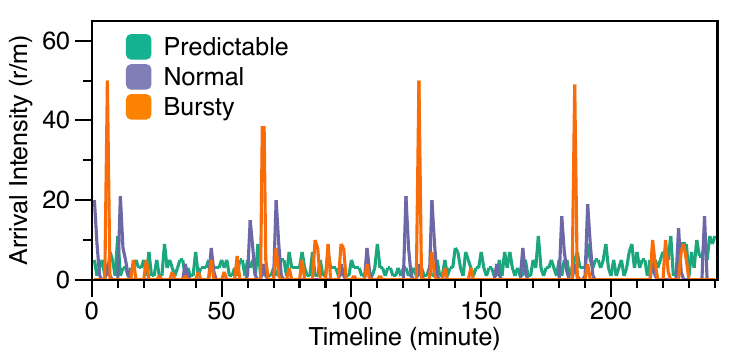} 
\caption{Example arrival traces for Predictable (CoV $\leq 1$), Normal ($1 <$ CoV $\leq 4$), and Bursty (CoV $> 4$) workloads.} 
\label{fig:workload_pattern}

\vspace{4pt}
\end{figure}

\noindent\textbf{Workload.} Following dLoRA~\cite{wu2024dlora} and HydraServe~\cite{lou2026hydraserve}, we leverage Microsoft Azure Function Trace~\cite{shahrad2020serverless} to generate workloads. Each LoRA variant is independently mapped to a randomly selected function trace. We classify traces by the coefficient of variation (CoV) of inter-arrival times: Predictable (CoV $\leq 1$), Normal ($1 <$ CoV $\leq 4$), and Bursty (CoV $> 4$), with examples in Fig.~\ref{fig:workload_pattern}.

\begin{figure*}[t]
\centering
\includegraphics[width=\textwidth]{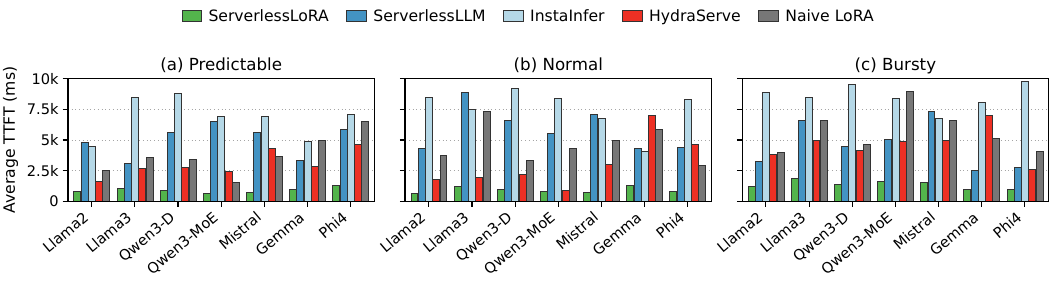}
\caption{Average \TTFT across model families under three arrival patterns. Qwen3-D denotes Qwen3-Dense.}
\label{fig:ttft_overall}
\vspace{4pt}
\end{figure*}

\begin{figure*}[t]
\centering
\includegraphics[width=\textwidth]{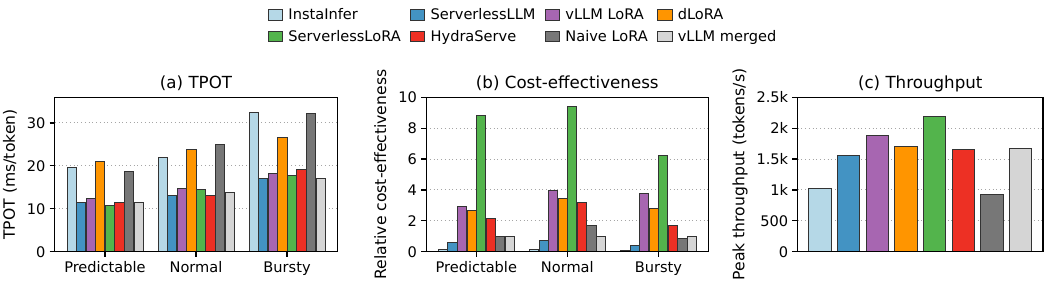}
\caption{(a) Average TPOT. (b) Cost-effectiveness relative to vLLM merged (=1). Both use three arrival patterns. (c) Peak output-token throughput.}
\label{fig:performance_metrics}
\label{fig:TPOT}
\label{fig:performance_overall}
\label{fig:throughput}
\vspace{4pt}
\end{figure*}

\noindent\textbf{Models and Libraries.} We evaluate 17 dense and mixture-of-experts (MoE) backbones from widely adopted model families (Table~\ref{tab:eval_models}), each with 10 popular LoRA adapters selected by Hugging Face download trends, totaling 170 LoRA variants. All inference pipelines use PyTorch and Transformers.

\noindent\textbf{Datasets and baselines.} Requests use GSM8K~\cite{cobbe2021gsm8k} prompts for mathematical reasoning and MMLU~\cite{hendrycks2021measuring} prompts for multi-task language understanding. We compare \sys with seven baselines. InstaInfer~\cite{sui2024pre}, ServerlessLLM~\cite{fu2024serverlessllm}, and HydraServe~\cite{lou2026hydraserve} reduce serverless cold starts through artifact preloading, accelerated checkpoint loading, and proactive model distribution with overlapped initialization, respectively. Naive\_LoRA shares backbone weight tensors via CUDA IPC (\S\ref{subsec:backbone_sharing}), without \sys's batched backbone execution, artifact warming, or contention-aware scheduling. Serverful baselines are dLoRA~\cite{wu2024dlora} and two configurations of vLLM v0.29.0~\cite{kwon2023efficient}: \emph{vLLM merged} deploys a separate merged model for each adapter, while \emph{vLLM LoRA} uses native multi-LoRA serving with multi-GPU support. All systems support batching; those without backbone sharing use merged inference to accelerate generation.

\noindent\textbf{Evaluation Metrics.} We evaluate the following metrics:
\begin{itemize}[leftmargin=*, topsep=3pt, itemsep=2pt, parsep=0pt]
    \item \textbf{Latency.} Cold-start latency covers pre-inference time, including container initialization and LLM artifact loading. TTFT measures the time from function invocation to the first token; TPOT is the mean interval between generated tokens.
    \item \textbf{Throughput.} We measure output tokens and served requests per unit time.
    \item \textbf{Cost and cost-effectiveness.} Monetary cost is the total workload cost. Cost-effectiveness is
    \[
        \frac{1}{\mathrm{E2E\_latency}\times\mathrm{Monetary\_Cost}}
    \]
    capturing the latency--cost trade-off: minimizing latency risks GPU over-provisioning, whereas minimizing cost increases cold starts.
\end{itemize}
We also evaluate \textbf{scalability} (latency/cost efficiency as workloads or resources scale) and \textbf{overhead} (\sys's additional latency and resource cost).

\noindent\textbf{Pricing model.} We estimate serverless costs using Azure Container Apps serverless GPU pricing~\cite{azure_container_apps_pricing}, accounting for all functions, including the Backbone Function. Serverful costs use Azure GPU VM pay-as-you-go pricing~\cite{azure_gpu_vm_pricing}. Resource allocations are based on each function's profiled peak-load requirements.

\begin{table}[t]
\centering
\small

\caption{Backbone LLMs used in our evaluation.}
\label{tab:eval_models}
\begin{tabularx}{\columnwidth}{@{}lX@{}}
\toprule
\textbf{Model Family} & \textbf{Backbones} \\
\midrule
Llama2~\cite{touvron2023llama}      & Llama2-7B, Llama2-13B, Llama2-70B \\
Llama3~\cite{grattafiori2024llama}  & Llama3-8B, Llama3.1-8B, Llama3-70B \\
Qwen3 Dense~\cite{yang2025qwen3}     & Qwen3-4B, Qwen3-8B, Qwen3-14B, Qwen3-32B \\
Qwen3 MoE~\cite{yang2025qwen3}       & Qwen3-30B-A3B \\
Mistral~\cite{jiang2023mistral,jiang2024mixtral}  & Mistral-7B, Mixtral-8x7B \\
Gemma~\cite{team2024gemma}           & Gemma-2-9B, Gemma-2-27B \\
Phi~\cite{abdin2024phi4}             & Phi-4-Mini, Phi-4 \\
\bottomrule
\end{tabularx}
\end{table}

\subsection{\TTFT and \TPOT Evaluation}
\label{subsection:TTFT}

\textbf{\TTFT.} Fig.~\ref{fig:ttft_overall} compares \sys with the four serverless baselines under Predictable, Normal, and Bursty workloads. \sys achieves the lowest \TTFT in every model-family/workload combination, reducing \TTFT by up to 92.3\% relative to InstaInfer.

ServerlessLLM and HydraServe accelerate model loading, while InstaInfer preloads model artifacts. These systems still load a complete backbone for each function. Naive\_LoRA shares the backbone weights but leaves variant-state preparation and cross-function execution uncoordinated. \sys combines backbone sharing with artifact warming and contention-aware scheduling to reduce these costs.

\textbf{\TPOT.} Fig.~\ref{fig:TPOT}(a) shows that \sys reduces \TPOT by 42.2--44.6\% relative to Naive\_LoRA across the three workloads. It remains close to vLLM merged and reduces \TPOT by 1.4--12.4\% relative to vLLM LoRA. It also outperforms InstaInfer and remains close to ServerlessLLM and HydraServe. On the Normal workload, its \TPOT is less than 10\% higher than that of ServerlessLLM and HydraServe.

\begin{figure}[t]
\centering
\includegraphics[width=\linewidth]{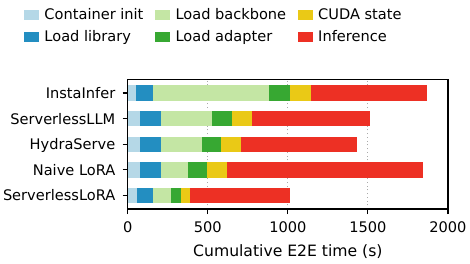}
\caption{Cumulative time breakdown for the Normal workload.}
\label{fig:breakdown}
\vspace{4pt}
\end{figure}

\begin{figure*}[t]
\centering
\includegraphics[width=\textwidth]{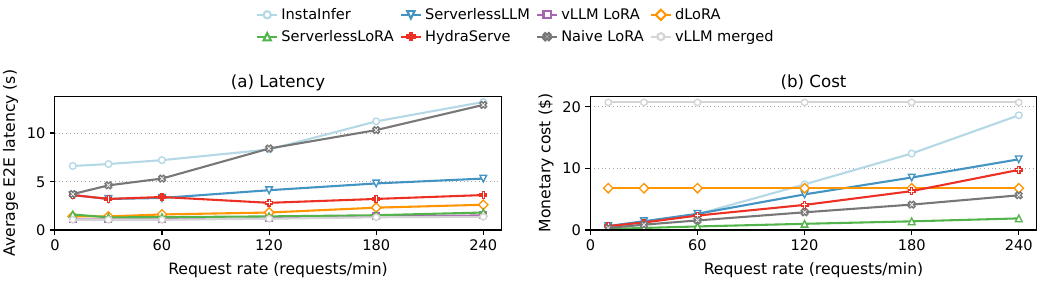}
\caption{Latency and monetary cost as the request rate increases from 10 to 240 requests/min.}
\label{fig:tradeofff}
\vspace{4pt}
\end{figure*}

\textbf{Tail latency.} Relative to the best serverless baseline for each metric and workload, \sys reduces P99 TTFT by 84.5--88.5\% and P99 TPOT by 16.3--25.1\% (Table~\ref{tab:tail_latency}). Together, these results show that \sys combines substantially lower TTFT with competitive mean TPOT and improved tail latency.

\begin{table}[!htbp]
\centering
\small
\captionsetup{skip=3pt}
\caption{P99 latency under three arrival patterns.}
\label{tab:tail_latency}
\setlength{\tabcolsep}{4pt}
\renewcommand{\arraystretch}{0.95}
\begin{tabular*}{\linewidth}[t]{@{\extracolsep{\fill}}lrrr@{}}
\toprule
(a) P99 TTFT (s) & Pred. & Normal & Bursty \\
\midrule
InstaInfer    & 248.3 & 233.6 & 195.3 \\
ServerlessLLM & 119.0 & 115.0 & 111.2 \\
HydraServe    & 219.7 & 253.0 & 190.5 \\
Naive\_LoRA   & 155.4 & 120.3 & 158.4 \\
\textbf{\sys} & \textbf{18.5} & \textbf{13.3} & \textbf{12.8} \\
\midrule
(b) P99 TPOT (ms/token) & Pred. & Normal & Bursty \\
\midrule
vLLM merged  & 21.5 & 20.8 & 20.6 \\
dLoRA        & 58.9 & 65.1 & 71.2 \\
InstaInfer   & 70.0 & 84.5 & 78.8 \\
ServerlessLLM & 29.5 & 33.9 & 28.9 \\
HydraServe   & 44.3 & 31.9 & 35.6 \\
Naive\_LoRA  & 72.6 & 82.2 & 73.4 \\
\textbf{\sys} & \textbf{23.3} & \textbf{23.9} & \textbf{24.2} \\
\bottomrule
\end{tabular*}
\end{table}

\subsection{Time Breakdown Analysis}

Fig.~\ref{fig:breakdown} shows that \sys reduces both cold-start work and inference time on the Normal workload. Inference accounts for most of \sys's cumulative time, whereas cold-start work still exceeds inference time in InstaInfer and ServerlessLLM. Naive\_LoRA also reduces backbone-loading time, but its inference time remains nearly twice that of \sys. Weight sharing alone therefore leaves substantial execution overhead.

\subsection{Cost-Effectiveness}

Fig.~\ref{fig:performance_overall}(b) normalizes cost-effectiveness to vLLM merged. Across the three workloads, \sys achieves 6.25--9.38$\times$ the cost-effectiveness of vLLM merged, 1.66--3.01$\times$ that of vLLM LoRA, and 2.20--3.26$\times$ that of dLoRA, while also outperforming all serverless baselines. Backbone reuse, warming, and coordinated execution reduce the latency--cost product.

\begin{figure}[t]
\centering
\includegraphics[width=\linewidth]{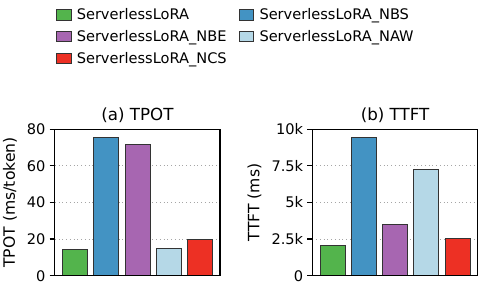}
\caption{Ablation of Backbone Sharing (NBS), batched backbone execution (NBE), Artifact Warming (NAW), and Contention-aware Scheduling (NCS).}
\label{fig:ablation}
\vspace{4pt}
\end{figure}

\subsection{Trade-off Between Latency and Cost}
\label{subsec:tradeoff}

We vary request rates while preserving arrival patterns. Fig.~\ref{fig:tradeofff} shows that \sys has the lowest cost at every rate and lower E2E latency than dLoRA except at the lowest rate. Averaged equally over the six rates, \sys reduces cost by 86.9\% and latency by 19.5\% relative to dLoRA. Against vLLM merged, mean latency increases by 27.5\% while mean cost falls by 95.7\%; against vLLM LoRA, mean latency increases by 20.8\% while mean cost falls by 86.9\%. vLLM LoRA and dLoRA share a cost curve because their resource configurations are identical.

\subsection{Throughput}
\label{subsection:throughput}

We evaluate peak output-token throughput for Llama-series functions. \sys achieves the highest throughput among all evaluated systems, reaching 2187 tokens/s (Fig.~\ref{fig:throughput}(c)). This is 2.37$\times$ the throughput of Naive\_LoRA, 31.1\% higher than vLLM merged, and 16.2\% higher than vLLM LoRA.

Backbone sharing leaves more GPU memory available for KV caches, while batched backbone execution combines work across isolated functions. The gap between \sys and Naive\_LoRA shows that sharing weights without coordinating execution does not realize the same throughput benefit.

\subsection{Ablation Study}

We compare \sys with four variants: \textbf{NBS}, without Backbone Sharing (\S\ref{subsec:backbone_sharing}); \textbf{NBE}, without batched backbone execution but retaining weight sharing; \textbf{NAW}, without Artifact Warming (\S\ref{subsec:warming}); and \textbf{NCS}, without Contention-aware Scheduling (\S\ref{subsec:scheduling_design}). We run a 4-hour Normal workload.

Fig.~\ref{fig:ablation} shows that Backbone Sharing benefits both startup and execution: disabling it increases \TPOT and \TTFT by factors of 5.24 and 4.60, respectively. NBE increases \TPOT 4.95$\times$ and \TTFT 1.70$\times$, showing that weight sharing alone leaves substantial execution overhead. Artifact Warming primarily reduces startup delay; disabling it increases \TTFT 3.52$\times$ while leaving \TPOT largely unchanged. Contention-aware Scheduling improves both metrics, with its removal increasing \TPOT by 35.1\% and \TTFT by 22.8\%.

\begin{figure}[t]
\centering
\includegraphics[width=\linewidth]{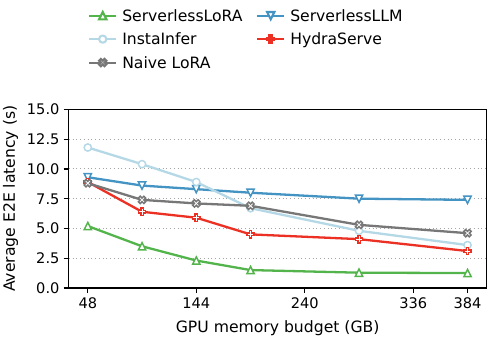}
\caption{E2E latency under increasing GPU memory budgets.}
\label{fig:scalability}
\vspace{10pt}
\end{figure}

\subsection{Scalability}

Fig.~\ref{fig:scalability} evaluates the workload containing all Qwen3-series functions as the GPU memory budget increases from 48 to 384~GB. \sys remains faster than all serverless baselines throughout, reducing E2E latency from 5.20 to 1.25~s. Gains diminish beyond 192~GB, with additional memory yielding only small latency improvements.

\subsection{Sensitivity Analysis}

We vary one parameter at a time on the Normal workload
(6,574 requests; Table~\ref{tab:sensitivity}).
Warming hit rate is the fraction of requests with complete
execution state ready at arrival, including state retained
after on-demand loading.
Hit rates vary little with either window; a 50\% memory
budget already achieves 98.39\%, with diminishing gains
thereafter.

\begin{table}[t]
  \centering
  \caption{Parameter sensitivity on the Normal workload.}
  \label{tab:sensitivity}
  \renewcommand{\arraystretch}{0.95}
  \small
  \setlength{\tabcolsep}{3pt}
  \begin{tabular*}{\columnwidth}
    {@{\extracolsep{\fill}}lrrrrrr@{}}
    \toprule
    Look-ahead $T$ (s)
      & 1 & 2 & 5 & 10 & 30 & 60 \\
    Hit rate (\%)
      & 98.45 & 98.39 & 98.39 & 98.39 & 98.39 & 98.39 \\
    \midrule
    History $W$ (s)
      & 5 & 15 & 30 & 60 & 120 & 300 \\
    Hit rate (\%)
      & 97.79 & 98.11 & 98.25 & 98.39 & 98.25 & 98.25 \\
    \midrule
    Idle budget (\%)
      & 10 & 25 & 50 & 75 & 100 & \\
    Hit rate (\%)
      & 55.64 & 76.71 & 98.39 & 99.38 & 99.51 & \\
    \bottomrule
  \end{tabular*}
  \par
  \parbox{\columnwidth}{\footnotesize
    CPU/GPU idle budgets scale together.
  }
\end{table}

\subsection{{Artifact-Warming Scheduling}}
\label{subsec:scheduling}

Since artifact pre-warming is an NP-hard PCKP, \sys uses the value-density heuristic in \S\ref{subsec:warming}. Simulations with 200 LoRA adapters, 20 backbone LLMs, 40 idle containers, and 80 GPUs show that the heuristic takes 0.04~s versus 10.6~s for the optimal solver, with total expected pre-warming benefit 1.5\% below the optimum.

\subsection{Overhead}

We evaluate a normal workload across all Qwen3 and Llama3 models, with frequently changing dynamic LoRA-backbone function bindings. Compared with vLLM LoRA, \sys incurs TTFT and TPOT overheads below 0.2~ms and 0.05~ms/token for Qwen3-8B, and below 1.2~ms and 0.2~ms/token for tensor-parallel Llama3-70B, respectively. Additional memory usage is 0.48~GB per GPU and 0.6~GB of CPU memory for the entire runtime.

\section{Related Work}
\label{sec:related}

\textbf{LLM serving:} Prior work accelerates LLM inference through iteration-level and chunked-prefill batching~\cite{yu2022orca,agrawal2023sarathi}, efficient attention and KV-cache management~\cite{dao2022flashattention,kwon2023efficient,lee2024infinigen,qin2025mooncake}, speculative decoding~\cite{miao2024specinfer}, and model parallelism or offloading~\cite{aminabadi2022deepspeed,li2023alpaserve,sheng2023flexgen,alizadeh2024llm}. DistServe~\cite{zhong2024distserve} further separates prefill and decode resources. These optimizations are orthogonal to \sys's cross-function state management: accelerating a serving engine does not prevent backbone duplication or repeated variant-state preparation across isolated serverless functions.

\noindent\textbf{Multi-LoRA LLM inference:} Punica~\cite{chen2024punica}, S-LoRA~\cite{sheng2023s}, Toppings~\cite{li2025toppings}, and dLoRA~\cite{wu2024dlora} improve adapter batching, memory management, or execution efficiency around a shared backbone. Their cross-adapter batching depends on keeping the participating adapters' state within a common serving engine. Splitting that engine into isolated functions fragments batches and reintroduces independent backbone reads, even if the weights remain shared. \sys enables batched backbone execution across functions while keeping adapter weights, adapter computation, and KV caches function-local.

\noindent\textbf{Optimizing serverless inference:} Prior systems improve serverless inference through batching~\cite{ali2020batch,yang2022infless,zhang2019mark}, model loading~\cite{li2022tetris,Optimus,pei2023asyfunc}, and artifact preloading~\cite{sui2024pre}. ServerlessLLM~\cite{fu2024serverlessllm}, Medusa~\cite{zeng2025medusa}, HydraServe~\cite{lou2026hydraserve}, and DEEPSERVE~\cite{hu2025deepserve} reduce LLM cold starts, while BlitzScale~\cite{zhang2025blitzscale}, FaaScale~\cite{yu2026faascale}, and FlexPipe~\cite{lin2026flexpipe} accelerate replica scaling and pipeline reconfiguration. GPU-sharing systems such as Torpor~\cite{yu2025torpor}, Aegaeon~\cite{xiang2025aegaeon}, Prism~\cite{yu2026prism}, and Dilu~\cite{lv2025dilu} improve utilization through model swapping, runtime reuse, or elastic resource allocation. Tensor sharing~\cite{li2022tetris} and GPU-centric data passing~\cite{wu2026grouter} reduce storage duplication and inter-function data movement. These mechanisms manage state residency and GPU allocation, but do not combine backbone operations across isolated LoRA functions. Thus, shared weights and warm runtimes still leave repeated HBM reads during independent execution. \sys reduces those reads through batched backbone execution while coordinating shared residency and variant-state warming.

\section{Conclusion}
\label{sec:conclusion}
This paper presented \sys, a serverless inference system designed for efficient LoRA LLM serving. We addressed the backbone redundancy, overlooked artifact loading, and resource contention problems in serverless LoRA inference. \sys combines read-only weight sharing and batched backbone execution with artifact warming and state-aware runtime coordination. Evaluation on industrial traces shows that \sys reduces \TTFT by up to 92.3\% against serverless baselines and achieves 6.25--9.38$\times$ the cost-effectiveness of vLLM merged.

\bibliographystyle{plainurl}

\bibliography{references}

\end{document}